\documentclass{article}

\usepackage{PRIMEarxiv}

\usepackage[utf8]{inputenc} 
\usepackage[T1]{fontenc}    
\usepackage{hyperref}       
\usepackage{url}            
\usepackage{booktabs}       
\usepackage{amsmath,amssymb,amsfonts}
\usepackage{algorithmic}
\usepackage{algorithm}
\usepackage{comment}
\usepackage{nicefrac}       
\usepackage{microtype}      
\usepackage{fancyhdr}       
\usepackage{graphicx}       
\graphicspath{{media/}}     
\usepackage{booktabs}

\title{An Integral Projection-based Semantic Autoencoder for Zero-Shot Learning
}

\author{
  William Heyden\thanks{Corresponding author: William Heyden (e-mail: william.heyden@nmbu.no)} ,
  {~~~~~~~~~~} Habib Ullah, 
  {~~~~~~~~~~} M. Salman Siddiqui, 
  {~~~~~~~~~~} Fadi Al Machot\\
  \\
  Faculty of Science and Technology (REALTEK) \\
  Norwegian University of Life Sciences \\ NMBU \\
  1430 Ås, Norway\\
}

\begin{document}
\maketitle

\begin{abstract}
Zero-shot Learning (ZSL) classification categorizes or predicts classes (labels) that are not included in the training set (unseen classes). Recent works proposed different semantic autoencoder (SAE) models where the encoder embeds a visual feature vector space into the semantic space and the decoder reconstructs the original visual feature space. The objective is to learn the embedding by leveraging a source data distribution, which can be applied effectively to a different but related target data distribution. Such embedding-based methods are prone to domain shift problems and are vulnerable to biases. We propose an integral projection-based semantic autoencoder (IP-SAE) where an encoder projects a visual feature space concatenated with the semantic space into a latent representation space. We force the decoder to reconstruct the visual-semantic data space. Due to this constraint, the visual-semantic projection function preserves the discriminatory data included inside the original visual feature space. The enriched projection forces a more precise reconstitution of the visual feature space invariant to the domain manifold. Consequently, the learned projection function is less domain-specific and alleviates the domain shift problem. Our proposed IP-SAE model consolidates a symmetric transformation function for embedding and projection, and thus, it provides transparency for interpreting generative applications in ZSL. Therefore, in addition to outperforming state-of-the-art methods considering four benchmark datasets, our analytical approach allows us to investigate distinct characteristics of generative-based methods in the unique context of zero-shot inference.
\end{abstract}

\maketitle

\section{Introduction}
\label{sec:introduction}
In a variety of studies, deep learning-based models have gained human-level abilities. However, these gains are conditional on the availability of high-quality and large-scale data. With the exponential growth of new classes in our real world, gathering enormous amounts of data is prohibitively costly and often infeasible. Additionally, annotating a sufficient amount of the data for training purposes for each class is resource intensive. As a consequence, several learning paradigms based on sparsely labelled data have been proposed, including semi-supervised learning, life-long learning, and active learning. These paradigms, however, are limited in their capacity to investigate changes in a small collection of labelled data. 

To address these problems, researchers developed embedding-based zero-shot learning (ZSL) models \cite{b1, b2, b3, b4}. They considered pre-trained models to evaluate test data of classes that have not been seen during the training stage. These models typically learn a projection function from a feature space to a semantic embedding space (e.g. attribute space). However, such a projection function is only concerned with predicting the training (seen) class semantic representation (e.g. attribute prediction) or classification. When applied to test data, which in the context of zero-shot contains different classes (unseen), these models typically suffer from the domain shift problem. In addition, the embedding-based model's final classification is subordinate to the nearest neighbour (NN) algorithm in the transformed space. The hubness problem is an inherent property of high dimensional data affecting the distribution of occurrences in this projected embedding space \cite{b49}. 

As a solution to these challenges, generative-based approaches \cite{b44, b8, b61, b62} were proposed. By generating class samples from available semantic representations, the NN search is reconditioned into supervised classification. Generative-based models alleviate the domain shift bias by generating authentic prototypes of the disjoint domain \cite{b63}. In \cite{b5}\cite{b7} and \cite{b8, b9}, the researchers are using Generative Adversarial Networks (GANs) and Variational Autoencoders (VAEs), respectively to acquire prototypes of the unseen distribution. GANs are known to easily diverge due to their adversarial nature \cite{b12} and VAEs to create blurry output as a result of the Kullback–Leibler divergence between the data and the distribution \cite{b13}. This impacts the generative capacity of prototype creation in a separate domain. In general, they are also not invertible, making them less suitable for downstream inference and reconstruction of unknown distributions \cite{b75}.

Based on the well-known semantic autoencoder model \cite{b10} (SAE), we propose the IP-SAE model to project both the visual feature space and the semantic feature space into the latent representation manifold. In addition, we use this model to demonstrate the adaptability of generative-based models for zero-shot inference. Building on autoencoder architecture, the encoder and decoder in our proposed model are multi-modular and share parameters. The encoder maps data into the integrated visual-semantic manifold. The decoder performs reconstruction of the original visual-semantic features. This projection alleviates the domain shift problem since the reconstructed samples of unseen domains are generated from the domain-invariant manifold. Furthermore, by adopting suitable regularisation parameters for the transformation function, we mitigate the hubness problem. We show that the generative abilities for zero-shot are a trade-off between stable transferability interpolating domains and performance in specific disjoint spaces. Our overall contributions can be summarised as follows:
\begin{itemize}
    \item  The proposed IP-SAE model uses an analytical solution to tackle the problem of ZSL. It has only one hyperparameter that should be tuned to increase the overall performance. Consequently, our results are reproducible, and the code will be published on GitHub \footnote{https://github.com/william-heyden/IP-SAE/}.
    \item Our IP-SAE model distinguishes from the prior SAE model \cite{b10} by reconstructing the latent semantic manifold by enriching the input space and actively using this low-dimension semantic manifold as a regularisation for inference of samples. 
    \item Average per-class accuracy is frequently reported in zero-shot learning state-of-the-art \cite{b56}, and the obtained results using the IP-SAE model show very high performance. In contrast to the state-of-the-art, we additionally use precision and recall to evaluate the model's end-to-end performance regarding the generalization to unseen classes and distinguish positive instances in unseen domains.
    \item We propose a methodology to improve the performance of generative ZSL by first, augmenting the input space to encompass multi-modal features; second, engaging regularisation to establish a complete latent manifold and third, leveraging the multi-model embedding of the disjoint data distribution to produce higher quality samples of the unseen classes.
    
\end{itemize}
\begin{figure*}[t]
\centering
\includegraphics[width=10 cm]{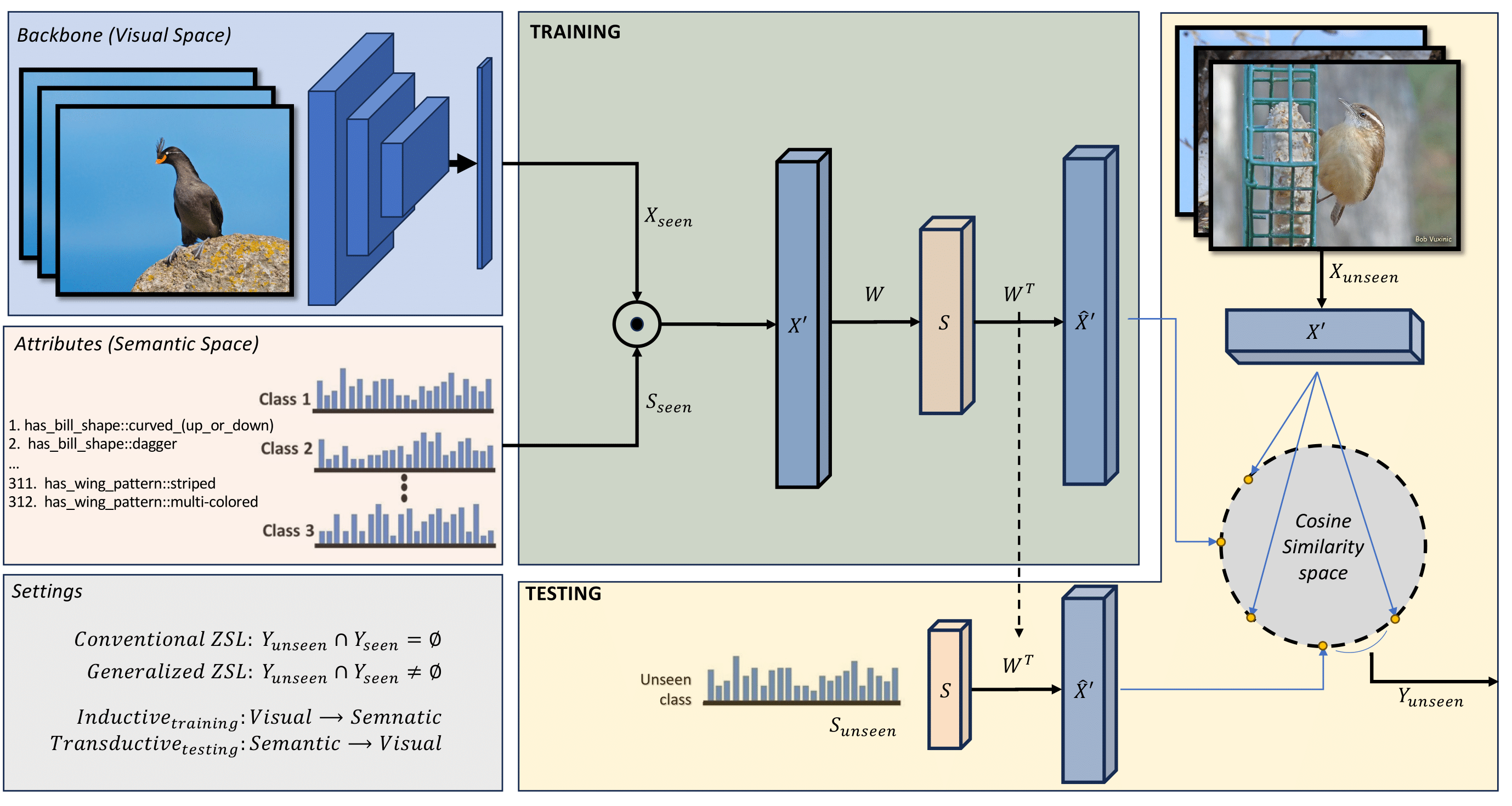}
\caption{The architecture of our proposed model. By enriching the input data space of the encoder, we are able to ensure surjectivity of the latent representational semantic space. As a result, we achieve enhanced quality and coverage of the reconstructed visual-feature space, leading to improved performance in nearest neighbor classification.}
\label{fig2}
\end{figure*}  

\section{Related works}
We classify the literature of zero-shot learning into two categories: embedding space-based  and feature generation-based.

\subsection{Embedding space-based methods}
Embedding space-based methods exploit a transfer function between semantic, visual, and latent feature space to close the gap between seen and unseen domains. Akata et al. \cite{b1} cultivated the relationships between features, attributes, and class embeddings. They were amongst the first to rank the compatibility of the visual image feature and the semantic feature embedding for correct classification, in contrast to prior work by researchers in \cite{b68} and \cite{b23} who used prediction of the class based on a learned mapping function.  Xian et al. \cite{b26} extended this work by introducing a collection of mapping functions ranked by a compatibility function between image and class embeddings. Their objective was to construct a piece-wise linear factorization of a non-linear compatibility function that learns the latent selection of components of the visual feature space.  \cite{b25, b27, b30} further explored different manifold structures to increase classification accuracy. The methods \cite{b14, b15, b16, b17} and \cite{b19} advanced the embedding function of the visual features and the semantic descriptors to a latent space. They proposed techniques to assist the transfer and the designs of appropriate embedding manifolds. Their novelty was to compose a discriminatory representation space that aligns the semantic space with the structure of the visual manifold. These models do not provide a natural mechanism for multiple modalities to be fused and optimized jointly in an end-to-end structure. In more recent work, \cite{b18} integrated a contrastive embedding model by learning a projection map from the embedding space and a comparative network to align the semantic and visual descriptors. While \cite{b70} used the shared label-space actively in training, to reconstruct the semantic representations. 

\subsection{Generative-based methods}
Generative-based methods work by generating pseudo data of the unseen domain to train a classifier impartial to both domain spaces. Reed et al. \cite{b69} predeveloped a GAN architecture effectively allowing for text or descriptive annotations to be translated into visual concepts. They briefly mention the zero-shot capabilities this implies. Mishra et al. \cite{b35} employed a similar approach, but expanded to the zero-shot setting exclusively. Using a VAE and conditioning on the semantic feature space, as opposed to the visual space, resulted in improved zero-shot inference. Li et al. \cite{b8} took advantage of the assumption that class representation originates from a prototypical space, encoding the relationship. This manifold structure is then learned from the data and used to generate synthetic observations. In Fadi et al. \cite{b80} the authors integrated two conditional autoencoders, of both modalities. The hybrid model generates pseudo training data from both decoders, which are then mapped to a final classifier. In a similar fashion, the researchers of the methods \cite{b34, b35, b36, b37} learned to consolidate the visual features for unseen classes using semantic information. These methods first learn a generative model considering variational autoencoder (VAE) and Generative adversarial networks (GAN) and then train a classifier using the complete space. With recent advancement made in generative algorithms, the performance of zero-shot architectures structured around generative-based methods has naturally also increased.

Li et al. \cite{b39} presented the Boomerang-GAN technique to find bilateral connections in zero-shot learning. They used a multimodal cycle-consistent loss to translate back the engendered features to semantic embeddings. Chou et al. \cite{b40} discovered the semantic-to-visual embeddings via a seamless fusion of adaptive and generative learning to investigate the correlation between image features and the corresponding semantic features. They stretched the semantic features of each class by supplementing image-adaptive attention so that the learned embedding could account for inter-class and intra-class variations. 

Xian et al. \cite{b46} combined VAE and GANs by assembling them into a conditional feature-generating model, called f-VAEGAN-D2, that synthesizes features from class embeddings. The authors \cite{b47} proposed the transformation and feedback-VAEGAN model (TF-VAEGAN). In addition to VAEs and GANs, they provided a semantic embedding decoder to reconstruct the embedding space. The decoder is used as a feedback module to improve the output of the Generator of the GAN. Both the f-VAEGAN-D2 model \cite{b46} and the TF-VAEGAN model \cite{b47} shows competitive performance. However, GANs and their derivatives show training instability, while VAE is more stable  \cite{b48}. 

\section{The proposed Approach}
\subsection{Problem definition}
The fundamental concept of ZSL is to construct a model that learns visual- and/or semantic cues translatable to unseen classes. In other words, generative zero-shot learning is required when all classes under observation lack labelled  training instances. As a result, the accessible dataset is divided into two groups: a training subset and a test subset. The training subset represented by seen classes $ \mathrm{Y}_{seen}=\{y^{1}_{seen}, y^{2}_{seen}, ..., y^{n}_{seen}\}$ and the test subset represented by unseen classes $ \mathrm{Y}_{unseen} =\{y^{1}_{unseen} , y^{2}_{unseen} , ..., y^{n}_{unseen} \}$. The assumption $ \mathrm{Y}_{seen} \cap  \mathrm{Y}_{unseen} = \phi$  should hold. In such a situation, the task is to build a model $\mathbb{R}^{d}  \rightarrow   \mathrm{Y}_{unseen}$ using only examples of training subsets to classify unseen classes. Afterwards, the trained classifier should be applied to test data of unseen classes under the zero-shot settings $\mathrm{Y}_{seen} \cap  \mathrm{Y}_{unseen} = \phi$. As a result, zero-shot learning offers a novel approach to overcome difficulties such as lack of training examples, with the goal of boosting a learning system's capacity to cope with unexpected situations in the same manner that individuals do.

To retain this similarity, most cutting-edge embedding-based solutions handle the ZSL issue by embedding the training data feature space and the semantic representation of class labels in some shared vector space. Unseen classes are then categorized according to a nearest-neighbour search. In the generalized zero-shot case, we seek to design a more generic model $ \mathbb{R}^{d}  \rightarrow  \mathrm{Y}_{seen} \cup \mathrm{Y}_{unseen} $, that can categorize/classify the seen and unseen classes appropriately. This implies that the test set contains data samples from both the seen and unseen classes \cite{b57}.

\subsection{Model}

We present a novel method to ZSL based on learning a Semantic AutoEncoder (SAE) inspired by \cite{b10}. The SAE method encodes the visual feature space of the training data into a semantic space. Their work is based on the assumption that a normal autoencoder is unsupervised leading to the fact that the latent space created by the learning process has no meaningful semantic representation. Therefore, they considered that each data point has a semantic representation and they forced the latent space to represent this semantic feature space. Taking advantage of the generative abilities of the autoencoder, the aim is then to learn visual feature prediction of the unseen classes. Classification of synthesised visual features is very challenging. Hence, the objective of the autoencoder generative space is to be close to the visual data space from unseen classes. The latent representation space is very limiting in its expressive power, given the complex distribution of image spaces. Our proposal implements an improved data space in a concatenation of visual- and semantic data space (fig. \ref{fig2}). This novelty helps to better extract class-discriminative components by increasing the expressive power of the latent representation space. We use a symmetric decoder of the encode-decode architecture to reconstruct an enriched visual-semantic sample space. This provides higher separability for unseen classes, accounting for the hubness problem effectively. We formulate the learning aim of the ZSL method as an optimization problem that minimizes the loss formulated as, 

\begin{equation}
\label{eq1}
\underset{W}{\text{minimize}}||X-W^TWX||^{2}_{F}~~~s.t. ~~ WX=S
\end{equation}

It represents the loss between the image visual space X and the semantic space S. F is the Frobenius norm and W is the weight. Solving equation (\ref{eq1}) with such a hard constraint $WX=S$ is not trivial. Therefore, the constraint can be relaxed into a softer one and the objective can be written as, 

\begin{equation}
\label{eq33}
\underset{W}{\text{minimize}} ||X-W^T S||^{2}_{F} + \lambda||WX - S||^{2}_{F}
\end{equation}

Where $\lambda$ is a weighting coefficient that controls the importance of the first and second terms. The symmetric terms correspond to the losses of the decoder and encoder, respectively. Equation (\ref{eq33}) has a quadratic form that can have a globally optimal solution. Therefore, taking the derivative of equation (\ref{eq33}) and setting it to zero, can lead to the analytical solution of this optimization problem \cite{b10}. The final solution can be formulated into the well-known Sylvester equation as, 

\begin{equation}
\label{eq444}
AW+WB = C;
\end{equation}

Where $A = SS^T, B = \lambda XX^T$, and $C = (1 + \lambda) SX^T$ such that $A$ and $B$ are positive semi-defined. Equation (\ref{eq444}) has an analytical solution that may be solved with efficiency using the Bartels-Stewart method \cite{b50}.
\par We propose that the visual representation space X can be replaced by a concatenated version consisting of both the semantic representation space S and the visual representation space X. It can be formulated as, 

\begin{equation}
\label{eq555}
X'= X \oplus S;
\end{equation}

To take into account the concatenation modelling, we reformulate equation (\ref{eq444}), where $A = SS^T, B = \lambda X'X'^T$, and $C = (1 + \lambda) SX'^T$. 
The importance of the concatenation is evident when we model the projection function (the decoder) as a linear ridge regression resulting in the formulation, 

\begin{equation}
\label{eq666}
\underset{W}{\text{minimize}}  ||X'-WS||^{2}_{F} + \lambda||W||^{2}_{F}
\end{equation}

The L2 norm calculates the distance of the vector coordinates from the origin of the vector space. It is well known that ridge regression has a closed-form solution $W = X'S^T(SS^T+\lambda I)^{-1}$. Thus, following the matrix norm properties: 

\begin{equation}
\label{eq777}
\begin{split}
||WS||_2 = ||X'S^T(SS^T+\lambda I)^{-1}S||_2 \leq\\ ||X'||_2||S^T(SS^T+\lambda I)^{-1}S||_2
\end{split}
\end{equation}

Using singular value decomposition (SVD), we can write,

\begin{equation}
\label{eq888}
||S^T(SS^T+\lambda I)^{-1}S||_2 = \frac{\alpha^2}{\alpha^2 + \lambda} \leq 1
\end{equation}

Where $\alpha$ is the largest singular value of S. So we have $||WS||_2 \leq ||X'||_2$. Consequently,  the mapped source data $||WS||_2$ is anticipated to be nearer to the origin of the space in relation to the target data $||X'||_2$ for the decoder, and vice versa for the encoder \cite{b28}. Therefore, this would consolidate the grouping of the data around the semantic space of a class and facilitate a clear separation between the classes. 
\par It is important to note that as the attributes of the semantic space (e.g., hand-annotated descriptive features) are sparse, matrix $S$ will be rank-deficient. The enhanced representation space will therefore, by the rank-nullity theorem of inequality $\dim (ker \mathrm{XS}) \leq  \dim (ker \mathrm{X}) + \dim(ker \mathrm{S}) $, result in the same being true for the visual space matrix $X'$. Therefore, applying Bartels-Stewart algorithm \cite{b50} to equation (\ref{eq444}) can no longer guarantee a unique solution, as at least $\mathrm{\textit{d}}-rank (\mathrm{S})$ with $d$ representing semantic dimensions, of the similarity matrices will be zero eigenvalues. The enriched space results in a higher-conditioned system sensitive to our choice of $\lambda$. This behaviour is evident in figure (\ref{fig:figure2}).

\subsubsection{Standard Zero-Shot Learning}
In the standard zero-shot setting our aim is to detect the classes of unseen data. The algorithm's output is the image's class label which is always an unseen class. The first step is to find $W$ which is the result of solving equation (\ref{eq444}). Then, we use $W^T$ to project prototypes of unseen enriched visual-semantic space from the latent representations. The final step for classification is then to calculate the cosine similarity between the projected visual space and the true visual space and label according to the most similar (top one) index label. Our approach handles the challenge of disjoint domain raised in the standard setting by enriching the input data space. The projection matrix obtained is therefore consistent with the semantic space provided, irrespective of the domain. Overfitting the data space is challenging for generative models in the convectional ZSL setting. Our proposed model solves this by estimating the lambda attaining optimal orthogonality in the encoded latent representation space. By trading off generative capabilities with information caption our model, and by extension generative models, are able to increase performance in this setting.

\subsubsection{Generalized Zero-Shot Learning}
\begin{figure}[t]
\footnotesize
\centering
\includegraphics[width=0.4\textwidth]{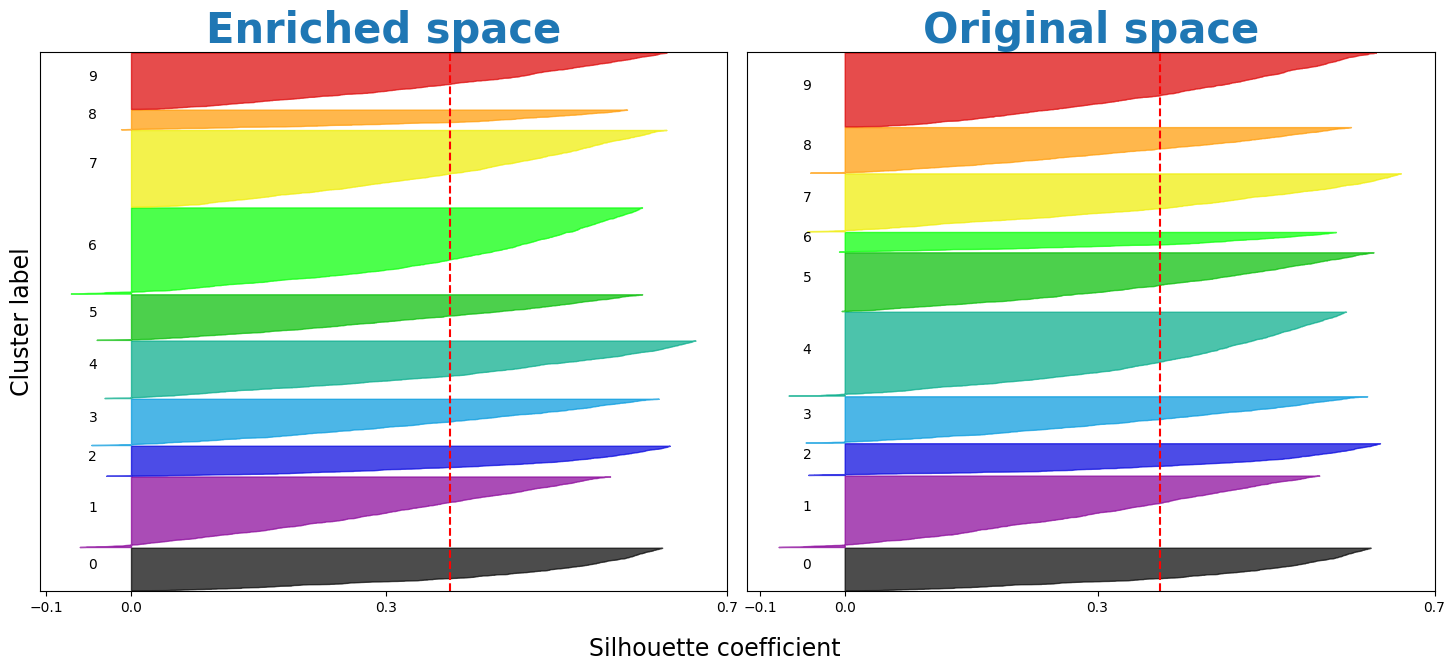}
\caption{Silhouette plotting the discriminative properties of enriched space of AwA2. "Enhanced space" reference our proposed visual-semantic space whereas "original space" corresponds to the visual feature space.}
\label{fig:figure3}
\end{figure}

In the more realistic generalized zero-shot setting (GZSL) we are presenting samples from both the seen and the unseen domain at testing $\mathrm{Y}_{seen} \cap  \mathrm{Y}_{unseen} \neq \phi$. 
We extract 20\% of data samples from the seen classes and mix them with the data samples from the unseen classes. In the generalized setting there is an inherent challenge of seen bias \cite{b54}, where the classes from the seen domain are significantly more represented in the final classification. Our approach alleviates this challenge also through a regularising lambda. The surjective properties of our proposed encoder ensure that the latent representation space is entirely mapped by every element of the visual feature space. Correcting for desired behaviour through lambda will align the information structure embedded in visual- and semantic- feature space. This structural alignment is directly transferable to the unseen domain, ensuring the quality and completeness of the synthetic visual-feature manifold. In generalized zero-shot learning the alignment of seen and unseen domain corrects for the domain shift issue in the projection space \cite{b6}. In our proposed enriched visual-semantic data space the structured projection space is further discriminated through increased class-wise distances. As theoretically proven in equation (\ref{eq888}), this is shown in figure (\ref{fig:figure3}) with the reduced intra-class space for each cluster (omitting class label 6 and 7).

\section{Experiments Analysis}
\label{exp}

\begin{figure*}[h]
\footnotesize
\centering
\includegraphics[width=12 cm]{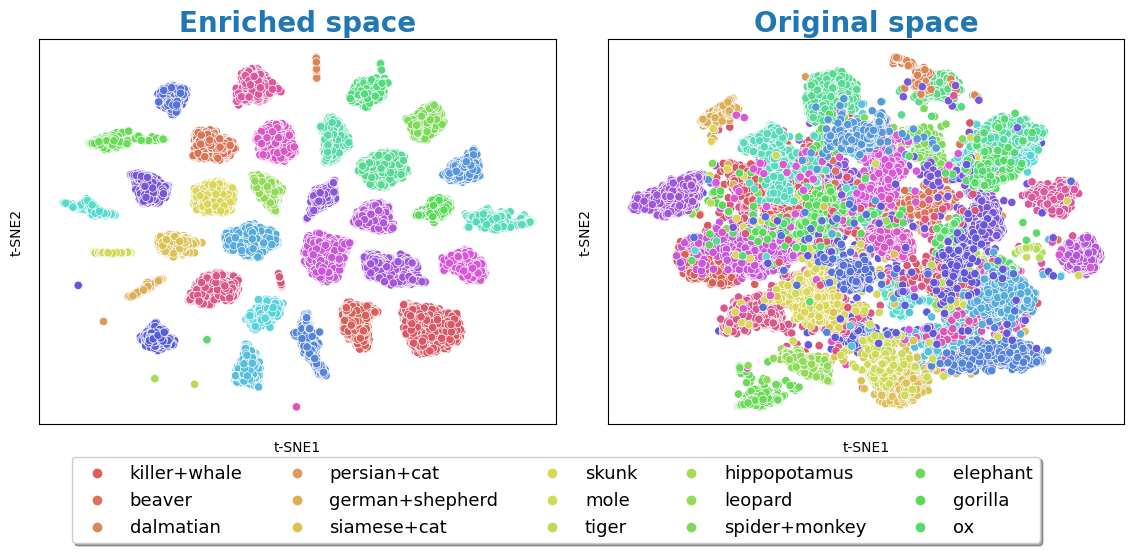}
\caption{Dataset visualization with a selection of labels. We present the visualization of the AwA2 dataset using T-SNE projection of features. The feature-semantic projection is shown on the left side and the semantic projection is shown on the right side. "Enhanced space" reference our proposed visual-semantic space, whereas "original space" corresponds to the visual feature space.}
\label{tsne}
\end{figure*} 

\subsection{Datasets}
To evaluate the performance of our proposed IP-SAE method, we consider four benchmark datasets. SUN Attribute (SUN)~\cite{sun} dataset, the CUB-200-2011 Bird (CUB) \cite{cub} dataset, and the AwA-1 and AwA-2 \cite{awa} datasets. The SUN dataset consists of 14340 images where 645 classes are seen and 72 are unseen. The AwA-1 dataset consists of 30475 images where 40 classes are seen and 10 are unseen. The AwA-2 dataset consists of 37322 images with 40 seen and 10 unseen classes. The CUB-200-2011 dataset consists of 11788 images where 150 classes are seen and 50 are unseen bird species. Each image additionally has 312 lower-level binary variables indicating visual properties (colour, pattern, form) of specific regions (beak, wings, tail, etc.). But attribute annotations are noisy. To denoise attributes, we used the concept bottleneck models \cite{b55}. We only counted the attributes as present if they were in at least 50\% of the images of the same class. Therefore, 200 lower-level features were chosen. 
\par In accordance with the norm of published research in zero-shot learning, we report the average per class top-1 accuracy to calculate the overall accuracy.

\begin{equation}
\label{eq4}
\text{acc}^{\text{\textit{per-class}}}_{\text{\textit{average}}} = \frac{1}{|Y|} \sum_{i=0}^{|Y|} \left(\frac{N^{\text{\textit{class}}_i}_{\text{\textit{correct\_class}}}}{N^{\text{\textit{class}}_i}_{\text{\textit{Total}}}}\right)
\end{equation}

In the conventional setting only the accuracy of unseen $class_i~\forall i \in Y_{unseen}$ while in the generative setting, we calculate the harmonic mean $H=\frac{2 \cdot A_{u} \cdot A_{s}}{A_{u}+A_{s}}$ between seen and unseen  $class_i~\forall i \in Y_{seen}$ \cite{b58}. In addition, we postulate for generative-based models in zero-shot classification the recall and precision are indispensable performance metrics, verifying a more nuanced evaluation of generative capabilities in the unseen domain.

\begin{equation}
    recall = \frac{TP}{TP+FN}, ~\text{ }~ precision = \frac{TP}{TP+FP}
\end{equation}

\subsection{Results}

For the visual space, we explored Resnet101 as the backbone architecture for the extracted features \cite{b56}. Concerning the semantic space, we rely on the semantic space vectors given by the authors of respective datasets. Regarding the GZSL, we looked at the empirical situation \cite{b57}. To eliminate the performance bias of the mapping, the nearest neighbour is selected using the cosine similarity. We apply equal regularisation value for the parameter $\lambda$ across all datasets to retain comparative behaviour. It can empirically be shown that by tuning the regulator to specific data manifolds a mutual orthogonal transformation matrix can be derived for the coarse dataset which will achieve near-perfect classification. 

Table \ref{tab:example11} illustrates the results under the conventional zero-shot setting, where the test data is disjoint from training. The results are reported with the suggested splits from \cite{b56}. Our proposed IP-SAE method outperformed the state-of-the-art methods by a high margin considering all four benchmark datasets.  In Table \ref{tab:example33}, we partition the results of our proposed method to consider the generalized zero-shot settings. Among existing methods, \cite{b56}, the classification accuracy of seen classes only is comparable with state-of-the-art methods across datasets. Considering unseen classes only, there is a significant improvement compared to similar methods for the fine-grained dataset CUB and SUN. The lower-level binary attributes depicting the visual characteristics of the data space are captured of higher quality in our proposed enriched visual-feature space. For the coarser annotated dataset AwA1 and AwA2, the performance is comparable to related embedding-based methods.

\begin{table}[h]
  \footnotesize
  \caption{The results of the disjoint assumption zero-shot setting in conjunction with the per-class accuracy measure.}
  \begin{center}
  \label{tab:example11}
  \begin{tabular}{|c|cccc|}
    \hline
\textbf{Model}	& \textbf{CUB}	 & \textbf{AWA1} & \textbf{AWA2}   & \textbf{SUN}\\
    \hline
DAP(PAMI'13) \cite{b17}		& 40.0			 	& 44.1	& 46.1& 39.9\\
IAP(PAMI'13) \cite{b17}			& 24.0			& 35.9	& 35.9 & 19.4\\
ConSE(arXiv'13) \cite{b22}		& 34.3		& 45.6	& 44.5&38.8\\
CMT(NeurIPS'13) \cite{b23}		& 34.6			& 39.5	& 37.9 & 39.9\\
SSE(arXiv'17) \cite{b65}		& 43.9		& 60.1	& 61.0 & 51.5\\
DeViSE(NeurIPS'13) \cite{b24}		& 52.0		& 54.2	& 59.7 & 56.5\\
SJE(CVPR'15) \cite{b25}		& 53.9		& 65.6	& 61.9 & 53.7\\
LATEM(CVPR'16) \cite{b26}		& 49.3		& 55.1	& 55.8 &55.3\\
ESZSL(ICML'15) \cite{b27}		& 53.9& 58.2	& 58.6 &54.5\\
ALE(PAMI'15)\cite{b1}		& 54.9		& 59.9	& 62.5 & 58.1\\
SYNC(CVPR'16) \cite{b3}		& 55.6	& 54.0	& 46.6 & 56.3\\
SAE(CVPR'17)\cite{b10}		& 33.3	& 53.0	& 54.1 & 40.3\\
Relation Net(CVPR'18)\cite{b32}		& 55.6	& 68.2	& 64.2 & -\\
DEM(CVPR'17)\cite{b28}		& 51.7		& 68.4	& 67.1& 61.9\\
f-VAEGAN-D2(CVPR'19)\cite{b46}		& 61.0	 & ----	& 71.1& 64.7\\
TF-VAEGAN(ECCV'20)\cite{b47}		& 64.9		& ---- 	& 72.2 & 66.0\\
CVAE(CVPR'18)\cite{b35}		& 52.1		 & 71.4	& 65.8& 61.7\\
GEM-ZSL(CVPR'21)\cite{b66} & 77.8		 & ----	& 67.3 & 62.8\\
AFRNet(AAAI'20)\cite{b67}& 50.3		 & 76.4	& 75.1 & 64.0\\
TransZero(AAAI'22)\cite{b77}& 76.8		 & ----	& 70.1 & 65.6\\
JG-ZSL(MDPI'23)\cite{b78}& 72.5		 & 70.6	& 69.4 & 60.3\\
HRT(ECCV'23)\cite{b79}& 71.7		 & ----	& 67.3 & 63.9\\
\hline
\textbf{ (Ours) }		& \textbf{80.1}		 & \textbf{92.9}  	& \textbf{82.0} & \textbf{94.4}\\
\hline
  \end{tabular}
  \end{center}
\end{table}

\begin{table}[h]
  \centering
  \footnotesize
  \caption{Results of Generalized Zero-Shot setting (GZSL) that are calculated based on the accuracy of seen classes, unseen classes, and~harmonic mean.}
  \footnotesize
  \begin{tabular}{|c|cccc|}
    \hline
\textbf{Dataset}	& \textbf{\%SeenClasses}	 & \textbf{Seen} & \textbf{Unseen}  & \textbf{Harmonic Mean}\\
    \hline
AWA1 & 20\% & 91.6 & 12.0 & 21.3\\
AWA2 & 20\% & 89.4  & 29.2 & 44.0\\
SUN &  20\% & 83.7  & 84.5 & 84.1\\
CUB &  20\% & 81.6  & 67.3& 73.7\\
    \hline
  \end{tabular}
  \label{tab:example33}
\end{table}

Furthermore, to highlight the impact of our proposed modelling, we present the data visualization in figure (\ref{tsne}). We use the t-SNE \cite{b59} approach to visually analyse the image feature vectors generated by our model for each class and compare them to the original image feature vectors for the AwA2 dataset. As can be seen, the original space (right) shows a separable feature space but high overlapping. In contrast to that, projecting the semantic space to the visual$-$semantic feature representation shows a clear separation between the classes (left). At the same time, the data will be grouped around the semantic space of a class which would overcome the problem of forming hubs. Consequently, data samples that belong to same class are well separable.

\begin{figure*}[t]
\centering
\footnotesize
\includegraphics[width=12 cm]{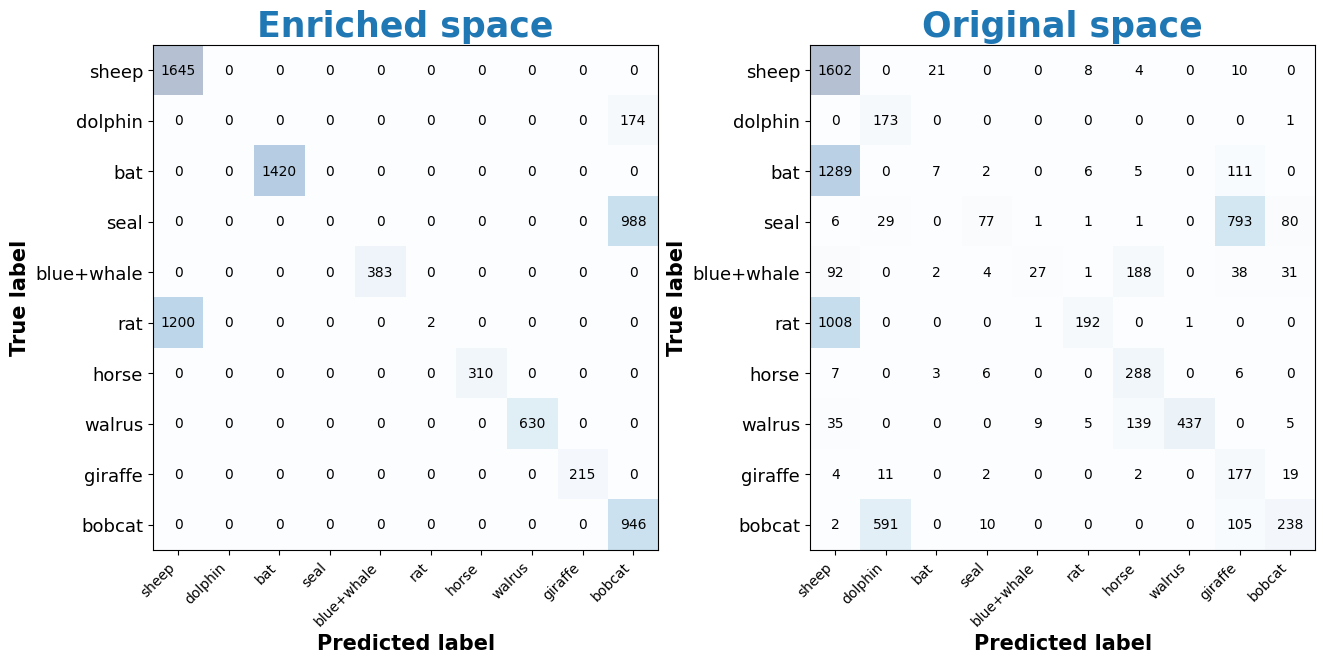}
\caption{{confusion} matrix of AWA2 dataset. Prediction of labels according to top-1 nearest neighbour. "Enhanced space" reference our proposed visual-semantic space, whereas "original space" corresponds to the visual feature space.}
\label{confusion}
\end{figure*}

The confusion matrix in figure (\ref{confusion}) reports a summary of the prediction of unseen classes in a matrix form \cite{b74}. We see a distinct main diagonal for our enriched visual-feature space (left) compared to the original space (right). This suggests that our proposed method is able to correctly identify and classify the true label from the projected latent representation space. Given the confusion matrix, we can extract the recall and precision measurements, as displayed in table \ref{tab:accuracy}. Here, we show that our proposed model has better precision and excels in classifying the true classes and correctly identifying false classes. This is universal for all benchmark datasets. This demonstrates that our IP-SAE method's capacity to generate a sample of high quality and the latent manifold is indeed expressive enough to cover the unseen distribution.

\section{Discussion}

\subsection{Generative properties}
\begin{figure}[h]
\footnotesize
\centering
\includegraphics[width=0.4\textwidth]{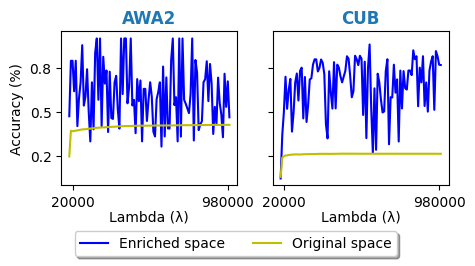}
\caption{Oscillating behaviour in accuracy given variation of lambdas. "Enriched space" reference our proposed visual-semantic space, whereas "original space" corresponds to the visual feature space.}
\label{fig:figure2}
\end{figure}

The generative aptness of autoencoders decreases when the latent representation space is overcomplete (e.g., higher dimension than the data space itself) \cite{b71}. Consequently, the composite function of the transformation matrix of eq. \ref{eq33}, $(W^T\circ W)$ the identity map of the visual space itself cannot be bijective. Implying we will never be able to recover the true visual feature space. It follows from functional analysis $W$ is a surjective transformation function in the encoder \cite{b51}. Suppose $f: V \rightarrow S$ such that $f$ is the transformation function that maps from the visual feature space $V$ to the semantic representation space $S$. In our enriched visual feature space, we can show that $\{f^{-1}(s)|s\in S\}$ is none empty. This implies that the columns of the transformation matrix $W$ are precisely the set of linear combinations to form a spanning set of the complete semantic representation space. This implication is not true for the original visual feature space, as the transformation matrix need not be full row rank. The encoder's surjective premise ensures that the enriched visual space image captures all available semantic information \cite{b51}. The effect of this is that we \textit{can} always construct a transformation function $g: S \rightarrow V$ satisfying $g(s)=v \in V~\forall s \in S$, meaning that $g \circ f$ is indeed an identity map. Hence, it is self-evident from the fact that an identity function is bijective \cite{b72} the decoder \textit{can} obtain injective properties. Therefore, the encoding ensures that enough information is preserved to recover the visual space, e.g., unique visual properties can be mapped distinctively from the semantic representation space and vice versa \cite{b52}. 
The analytical solution offered by the Sylvester equation (\ref{eq444}) finds the optimal transformation matrix between the visual and the semantic space. By enriching the visual space of the seen domain, we are condensing the distance to the subjective image of the unseen domain, hence improving the embedding quality of semantic space for the unseen domain.
This is a trade-off between information preservation of the encoding and the generative capabilities of the decoding. The oscillating behaviour seen in figure (\ref{fig:figure2}) occurs due to the reconstructed visual space having unique representations in the semantic representation space and multiple reconstructed depictions. 
However, by the properties of ridge regularisation and the elegant symmetry in the model, we can alleviate this challenge of high variance by designing the coefficients of the transformation matrix through a lambda. The objective is the optimal trade-off between the encoding and decoding of the available information in the seen domain.
For generative models, a low-dimension latent space achieves regulatory properties \cite{b73}. Since the seen- and unseen domain share semantic feature representation, the design of an elaborate mapping function or embedding space is critical for transferring knowledge between classes \cite{b6}. Our results show that the visual-semantic latent space span should be expressive by the mapping function, e.g., the mapping of available knowledge grants surjectivity. The limitation in our proposed model is that we force the same mapping also to be inversely injective, which is contradictory to autoencoders.

\subsection{Regularisation of $\lambda$ for ZLS}
The regularisation term of the loss function (\ref{eq33}) characterises with ridge regression. It reduces the condition number of the near singular similarity matrix $XX^T$ of equation (\ref{eq444}) by adding a non-negative element to the diagonalizable matrix. Increasing the singular values in the Schur decomposition (of the Bartels-Stewart algorithm) for the visual space reduces noise in the decoding due to the inverse proportional properties in the reversed operation \cite{b53}. The ridge behaviour of lambda ensures discriminating visual and semantic structures are preserved in the transformation matrix. Redundant projections of the high variance found in visual space are replaced with the compact semantic space allowing the projection matrix to learn the direction with the lowest variance by increasing the regularisation. Since we are only concerned with the relative position in the visual semantic feature space, an ambiguous loss of information is adequate in the recovery of the manifold by the decoder. We hypothesize that by optimising for $\lambda$, the projection matrix favours shared principle components of the visual and semantic space, a result of minimizing $W$ in the loss function (\ref{eq33}).

\begin{table}[h]
\centering
\footnotesize
\caption{Accuracy measurement for respective datasets in conventional zero-shot learning setting.}
\begin{tabular}{|l|lll|lll|}
\hline
          & \multicolumn{3}{c|}{\textit{Enriched}}      & \multicolumn{3}{c|}{\textit{Original}}       \\
          & \textbf{AWA1} & \textbf{CUB} & \textbf{SUN} & \textbf{AWA1} & \textbf{CUB} & \textbf{SUN} \\
          \hline
Precision & 0.6229        & 0.8032       & 0.9166       & 0.6044        & 0.3135       & 0.4805       \\
Recall    & 0.7359        & 0.8587       & 0.9444       & 0.4571        & 0.2150       & 0.3479       \\
F1        & 0.6568        & 0.8197       & 0.9259       & 0.3902        & 0.1924       & 0.3557\\
\hline
\end{tabular}
\label{tab:accuracy}
\end{table}

\subsection{Ablation Study}
Table \ref{tab:example33} evaluates the transferability of our elevated semantic representation space. The encoded representation space is the source of shared information for the seen and unseen domain. The efficiency of the transformation matrix $W$ introduces a compromise between $I$) A smooth embedding of the semantic manifold to leverage available domain (seen and unseen) information. $II$) A stable projection into the visual manifold for discriminative, high-quality sample distribution. This (dis)entanglement of the transformation matrix is due to the symmetric in our model (objective function \ref{eq33}). For the encoder, we see that for the finer-grained datasets SUN and CUB, the discrepancies in the accuracy of seen and unseen classes are less. For coarser datasets (AwA1 and AwA2), the model efficiently captures relevant information but struggles to exploit it. The lower harmonic mean shows this. In terms of generative abilities in a disjoint domain, this trivially implies that the reduced semantic information stimulated by increased regularisation of the surjective mapping prompts an advantage to injective properties of the projection.

Table \ref{tab:accuracy} shows the influence of an embedding function with a spanning set over the semantic representation space. For our enriched space $recall>precision$ while the original space results in $recall<precision$. This suggests a complete visual feature reproduction space. The generative transformation is not guaranteed to be expressive enough to replicate all unseen classes in the high-dimensional visual feature space. However, capturing transferable information between the disjoint domains shows to assist progressively in terms of its position in reproduced space and coverage of the manifold by the generator.
Note that overall recall and precision in our enriched space are still greater than in the original space.

\subsection{Performance Metrics}
In accordance with the progress made within zero-shot learning, there has been extensive research on comparing state-of-the-art algorithms \cite{b56, b63}. To evaluate performance and demonstrate abilities of proposed models, the average class accuracy, eq. (\ref{eq4}), have exclusively been included in publications. We argue that the classic definition of accuracy to report generative-based zero-shot learning effectiveness is insufficient. This only discloses the average of per class true positive predictions, which is biased towards sample size and favours less exact spatial arrangement of the representational manifold \cite{b76}. To comprehensively capture the impact of generative zero-shot learning, it is recommended to consider two crucial aspects: a) whether the manifold of knowledge transfer adequately covers the real distribution and b) if the quality of generated class prototypes reflects reality. Consequently, we advocate for adopting the definitions of recall and precision. By examining recall and precision, we can gain insights into the generative process and accurately interpret the model's performance.

\section{Broader Impact}
In the assessment of an analytically solvable algorithm designed for generative zero-shot learning this research show how enriching and regularising multi-modal spaces affects the generative capabilities of the disjoint domains. For zero-shot learning researchers, the conceptual insights derived can be applied to improve performance across various generative-based methodologies. In addition, this work can be a contrivance for improving any larger-scale recognition systems by reducing the dependency on the label of the data. At last, this work enables high-quality samples to be synthesised of unseen distribution, thereby essentially implicating a new aspect of creativity for generative models.

\section{Conclusions}
\label{conc}
In this study, we proposed IP-SAE zero-shot learning model. The proposed work included extensive testing and coverage of generative qualities for zero-shot projections. In the IP-SAE model on four benchmark datasets, our method outperformed the state-of-the-art methods using the precision, the recall, the f-score and the per-class accuracy evaluation metrics. In addition, it showed high performance in both, the conventional ZSL and the generalized zero-shot setting. We selected the analytical solution to show how generative approaches for zero-shot learning can be enhanced by; expanding the data manifold to ensure completeness of the disjoint domains; and by regularising the latent representation to augment the sample manifold.

In future work, we aim to improve the generalized zero-shot learning model by using a generative model on a project matrix of the visual and semantic features. This will improve the transformation function's orthogonality and generate samples into a new embedding space with more distinct classifications.


\newpage
\begin{thebibliography}{00}

\bibitem{b1} Z. Akata, F. Perronnin, Z. Harchaoui, C. Schmid, ''Label-embedding for image classification'' in IEEE transactions on pattern analysis and machine intelligence, \emph{IEEE,} volume 38, n. 7, 2015, pp. 1425-138

\bibitem{b2} Z. Akata, S. Reed, D Walter, H Lee, B. Schiele, ''Evaluation of output embeddings for fine-grained image classification'', in Proceedings of the IEEE conference on computer vision and pattern recognition, 2015, pp. 2927--2936

\bibitem{b3} S. Changpinyo, W Chao, B Gong, F Sha,  ''Synthesized classifiers for zero-shot learning'' in Proceedings of the IEEE conference on computer vision and pattern recognition, 2016, pp. 5327-5336

\bibitem{b4} G. Xie,  L. Liu, F. Zhu, F. Zhao, Z. Zhang, Y. Yao, J. Qin, S. Jiem, L. Shao, ''Region graph embedding network for zero-shot learning'' in European conference on computer vision, \emph{Springer,} 2020, pp. 562-580

\bibitem{b5} X. Zhao, Y. Shen, S. Wang, H. Zhang,''Generating diverse augmented attributes for generalized zero-shot learning'' in \emph{Patter Recognition Letters}, Volume 166, 2023, pp. 126-133.

\bibitem{b6} Wang, C., Min, S., Chen, X., Sun, X. and Li, H., 2021. ''Dual progressive prototype network for generalized zero-shot learning'' Advances in Neural Information Processing Systems, 34, pp.2936-2948.

\bibitem{b7} Schonfeld, E., Ebrahimi, S., Sinha, S., Darrell, T. and Akata, Z., 2019. ''Generalized zero-and few-shot learning via aligned variational autoencoders.'' In Proceedings of the IEEE/CVF Conference on Computer Vision and Pattern Recognition (pp. 8247-8255).

\bibitem{b8} Li, Y. and Wang, D., 2017. ''Zero-shot learning with generative latent prototype model.'' arXiv preprint arXiv:1705.09474.

\bibitem{b9} Mishra, A., Krishna Reddy, S., Mittal, A. and Murthy, H.A., 2018. ''A generative model for zero shot learning using conditional variational autoencoders.'' In Proceedings of the IEEE conference on computer vision and pattern recognition workshops (pp. 2188-2196).

\bibitem{b10} Kodirov, E., Xiang, T. and Gong, S., 2017. ''Semantic autoencoder for zero-shot learning.'' In Proceedings of the IEEE conference on computer vision and pattern recognition (pp. 3174-3183).

\bibitem{b11} Pourpanah, F., Abdar, M., Luo, Y., Zhou, X., Wang, R., Lim, C.P., Wang, X.Z. and Wu, Q.J., 2022. ''A review of generalized zero-shot learning methods.'' IEEE transactions on pattern analysis and machine intelligence.

\bibitem{b12} Lee, M. and Seok, J., 2020. ''Regularization methods for generative adversarial networks: An overview of recent studies.'' arXiv preprint arXiv:2005.09165.

\bibitem{b13} Diederik P. Kingma, Max Welling ''An Introduction to Variational Autoencoders'' in \emph{Foundations and Trends in Machine Learning} Vol. 14, No. 4, 2019, pp 1–18.

\bibitem{b14} Huynh, D. and Elhamifar, E., 2020. ''Fine-grained generalized zero-shot learning via dense attribute-based attention.'' In Proceedings of the IEEE/CVF conference on computer vision and pattern recognition (pp. 4483-4493).

\bibitem{b15} Xie, G.S., Liu, L., Jin, X., Zhu, F., Zhang, Z., Qin, J., Yao, Y. and Shao, L., 2019. ''Attentive region embedding network for zero-shot learning.'' In Proceedings of the IEEE/CVF conference on computer vision and pattern recognition (pp. 9384-9393).

\bibitem{b16} Min, S., Yao, H., Xie, H., Wang, C., Zha, Z.J. and Zhang, Y., 2020. ''Domain-aware visual bias eliminating for generalized zero-shot learning.'' In Proceedings of the IEEE/CVF Conference on Computer Vision and Pattern Recognition (pp. 12664-12673).

\bibitem{b17} Lampert, C.H., Nickisch, H. and Harmeling, S., 2013. ''Attribute-based classification for zero-shot visual object categorization''. IEEE transactions on pattern analysis and machine intelligence, 36(3), pp.453-465.

\bibitem{b18} Han, Z., Fu, Z., Chen, S. and Yang, J., 2021. ''Contrastive embedding for generalized zero-shot learning''. In Proceedings of the IEEE/CVF Conference on Computer Vision and Pattern Recognition (pp. 2371-2381).

\bibitem{b19} Fu, Y., Hospedales, T.M., Xiang, T., Fu, Z. and Gong, S., 2014. ''Transductive multi-view embedding for zero-shot recognition and annotation''. In Computer Vision–ECCV 2014: 13th European Conference, Zurich, Switzerland, September 6-12, 2014, Proceedings, Part II 13 (pp. 584-599). Springer International Publishing.

\bibitem{b20} Kodirov, E., Xiang, T., Fu, Z. and Gong, S., 2015. ''Unsupervised domain adaptation for zero-shot learning''. In Proceedings of the IEEE international conference on computer vision (pp. 2452-2460).

\bibitem{b21} Zhang, J., Li, Q., Geng, Y.A., Wang, W., Sun, W., Shi, C. and Ding, Z., 2022. ''A zero-shot learning framework via cluster-prototype matching''. Pattern Recognition, 124, p.108469.

\bibitem{b22} Norouzi, M., Mikolov, T., Bengio, S., Singer, Y., Shlens, J., Frome, A., Corrado, G.S. and Dean, J., 2013. ''Zero-shot learning by convex combination of semantic embeddings''. arXiv preprint arXiv:1312.5650.

\bibitem{b23} Socher, R., Ganjoo, M., Manning, C.D. and Ng, A., 2013. ''Zero-shot learning through cross-modal transfer''. Advances in neural information processing systems, 26.

\bibitem{b24} Frome, A., Corrado, G.S., Shlens, J., Bengio, S., Dean, J., Ranzato, M.A. and Mikolov, T., 2013. ''Devise: A deep visual-semantic embedding model''. Advances in neural information processing systems, 26.

\bibitem{b25} Akata, Z., Reed, S., Walter, D., Lee, H. and Schiele, B., 2015. ''Evaluation of output embeddings for fine-grained image classification''. In Proceedings of the IEEE conference on computer vision and pattern recognition (pp. 2927-2936).

\bibitem{b26} Xian, Y., Akata, Z., Sharma, G., Nguyen, Q., Hein, M. and Schiele, B., 2016. ''Latent embeddings for zero-shot classification''. In Proceedings of the IEEE conference on computer vision and pattern recognition (pp. 69-77).

\bibitem{b27} Romera-Paredes, B. and Torr, P., 2015, June. ''An embarrassingly simple approach to zero-shot learning''. In International conference on machine learning (pp. 2152-2161). PMLR.

\bibitem{b28} Zhang, L., Xiang, T. and Gong, S., 2017. ''Learning a deep embedding model for zero-shot learning''. In Proceedings of the IEEE conference on computer vision and pattern recognition (pp. 2021-2030).

\bibitem{b29} Li, Q., Hou, M., Lai, H. and Yang, M., 2022. ''Cross-modal distribution alignment embedding network for generalized zero-shot learning''. Neural Networks, 148, pp.176-182.

\bibitem{b30} Annadani, Y. and Biswas, S., 2018. ''Preserving semantic relations for zero-shot learning''. In Proceedings of the IEEE Conference on Computer Vision and Pattern Recognition (pp. 7603-7612).

\bibitem{b31} Yu, Y., Ji, Z., Han, J. and Zhang, Z., 2020. ''Episode-based prototype generating network for zero-shot learning''. In Proceedings of the IEEE/CVF conference on computer vision and pattern recognition (pp. 14035-14044).

\bibitem{b32} Sung, F., Yang, Y., Zhang, L., Xiang, T., Torr, P.H. and Hospedales, T.M., 2018. ''Learning to compare: Relation network for few-shot learning''. In Proceedings of the IEEE conference on computer vision and pattern recognition (pp. 1199-1208).

\bibitem{b33} Badirli, S., Akata, Z., Mohler, G., Picard, C. and Dundar, M.M., 2021. ''Fine-grained zero-shot learning with dna as side information''. Advances in Neural Information Processing Systems, 34, pp.19352-19362.

\bibitem{b34} Bucher, M., Herbin, S. and Jurie, F., 2017. ''Generating visual representations for zero-shot classification''. In Proceedings of the IEEE International Conference on Computer Vision Workshops (pp. 2666-2673).

\bibitem{b35} Mishra, A., Krishna Reddy, S., Mittal, A. and Murthy, H.A., 2018. ''A generative model for zero shot learning using conditional variational autoencoders''. In Proceedings of the IEEE conference on computer vision and pattern recognition workshops (pp. 2188-2196).

\bibitem{b36} Verma, V.K., Arora, G., Mishra, A. and Rai, P., 2018. ''Generalized zero-shot learning via synthesized examples''. In Proceedings of the IEEE conference on computer vision and pattern recognition (pp. 4281-4289).

\bibitem{b37} Xian, Y., Lorenz, T., Schiele, B. and Akata, Z., 2018. ''Feature generating networks for zero-shot learning''. In Proceedings of the IEEE conference on computer vision and pattern recognition (pp. 5542-5551).

\bibitem{b38} Xu, B., Zeng, Z., Lian, C. and Ding, Z., 2022. ''Generative mixup networks for zero-shot learning''. IEEE Transactions on Neural Networks and Learning Systems.

\bibitem{b39} Li, J., Jing, M., Lu, K., Zhu, L. and Shen, H.T., 2021. ''Investigating the bilateral connections in generative zero-shot learning''. IEEE Transactions on Cybernetics, 52(8), pp.8167-8178.

\bibitem{b40} Chou, Y.Y., Lin, H.T. and Liu, T.L., 2021. ''Adaptive and generative zero-shot learning''. In International conference on learning representations.

\bibitem{b41} Ma, P., Lu, H., Yang, B. and Ran, W., 2022. ''GAN-MVAE: A discriminative latent feature generation framework for generalized zero-shot learning''. Pattern Recognition Letters, 155, pp.77-83.

\bibitem{b42} Ye, Y., Pan, T., Luo, T., Li, J. and Shen, H.T., 2022. ''Learning Modality-Consistent Latent Representations for Generalized Zero-Shot Learning''. IEEE Transactions on Multimedia.

\bibitem{b43} Lu, Z., Lu, Z., Yu, Y. and Wang, Z., 2022. ''Learn more from less: Generalized zero-shot learning with severely limited labeled data''. Neurocomputing, 477, pp.25-35.

\bibitem{b44} Li, J., Jing, M., Lu, K., Ding, Z., Zhu, L. and Huang, Z., 2019. ''Leveraging the invariant side of generative zero-shot learning''. In Proceedings of the IEEE/CVF Conference on Computer Vision and Pattern Recognition (pp. 7402-7411).

\bibitem{b45} Vyas, M.R., Venkateswara, H. and Panchanathan, S., 2020. ''Leveraging seen and unseen semantic relationships for generative zero-shot learning''. In Computer Vision–ECCV 2020: 16th European Conference, Glasgow, UK, August 23–28, 2020, Proceedings, Part XXX 16 (pp. 70-86). Springer International Publishing.

\bibitem{b46} Xian, Y., Sharma, S., Schiele, B. and Akata, Z., 2019. ''f-vaegan-d2: A feature generating framework for any-shot learning''. In Proceedings of the IEEE/CVF conference on computer vision and pattern recognition (pp. 10275-10284).

\bibitem{b47} Narayan, S., Gupta, A., Khan, F.S., Snoek, C.G. and Shao, L., 2020. ''Latent embedding feedback and discriminative features for zero-shot classification''. In Computer Vision–ECCV 2020: 16th European Conference, Glasgow, UK, August 23–28, 2020, Proceedings, Part XXII 16 (pp. 479-495). Springer International Publishing.

\bibitem{b48} Zhang, T., Yang, Z. and Li, D., 2022. ''Stochastic simulation of deltas based on a concurrent multi-stage VAE-GAN model''. Journal of Hydrology, 607, p.127493.

\bibitem{b49} Radovanovic, M., Nanopoulos, A. and Ivanovic, M., 2010. ''Hubs in space: Popular nearest neighbors in high-dimensional data''. Journal of Machine Learning Research, 11(sept), pp.2487-2531.

\bibitem{b50} Bartels, R.H. and Stewart, G.W., 1972. ''Solution of the matrix equation AX+ XB= C [F4]''. Communications of the ACM, 15(9), pp.820-826.

\bibitem{b51} Bretscher, O., 1997. ''Linear algebra with applications (Vol. 52)''. Eaglewood Cliffs, NJ: Prentice Hall. Ch. 3

\bibitem{b52} Artin, M. 2011. ''Algebra''. Pearson Education. ISBN 9780132413770.  

\bibitem{b53} Strang, G., 2006. ''Linear algebra and its applications''. Belmont, CA: Thomson, Brooks/Cole. p260.

\bibitem{b54} Song, J., Shen, C., Yang, Y., Liu, Y. and Song, M., 2018. ''Transductive unbiased embedding for zero-shot learning''. In Proceedings of the IEEE conference on computer vision and pattern recognition (pp. 1024-1033).

\bibitem{sun} Patterson, G. and Hays, J., 2012, June. ''Sun attribute database: Discovering, annotating, and recognizing scene attributes''. In 2012 IEEE Conference on Computer Vision and Pattern Recognition (pp. 2751-2758). IEEE.

\bibitem{awa} Zhao, B., Fu, Y., Liang, R., Wu, J., Wang, Y. and Wang, Y., 2019. ''A large-scale attribute dataset for zero-shot learning''. In Proceedings of the IEEE/CVF Conference on Computer Vision and Pattern Recognition Workshops (pp. 0-0).

\bibitem{cub} Wah, C., Branson, S., Welinder, P., Perona, P. and Belongie, S., 2011. The caltech-ucsd birds-200-2011 dataset.

\bibitem{b55} Koh, P.W., Nguyen, T., Tang, Y.S., Mussmann, S., Pierson, E., Kim, B. and Liang, P., 2020, November. ''Concept bottleneck models''. In International Conference on Machine Learning (pp. 5338-5348). PMLR.

\bibitem{b56} Xian, Y., Schiele, B. and Akata, Z., 2017. ''Zero-shot learning-the good, the bad and the ugly''. In Proceedings of the IEEE conference on computer vision and pattern recognition (pp. 4582-4591).

\bibitem{b57} Chao, W.L., Changpinyo, S., Gong, B. and Sha, F., 2016. ''An empirical study and analysis of generalized zero-shot learning for object recognition in the wild''. In Computer Vision–ECCV 2016: 14th European Conference, Amsterdam, The Netherlands, October 11-14, 2016, Proceedings, Part II 14 (pp. 52-68). Springer International Publishing.

\bibitem{b58} Xian, Y., Lorenz, T., Schiele, B. and Akata, Z., 2018. ''Feature generating networks for zero-shot learning''. In Proceedings of the IEEE conference on computer vision and pattern recognition (pp. 5542-5551).

\bibitem{b59} Van der Maaten, L. and Hinton, G., 2008. ''Visualizing data using t-SNE''. Journal of machine learning research, 9(11).

\bibitem{b60} Wang, W., Pu, Y., Verma, V., Fan, K., Zhang, Y., Chen, C., Rai, P. and Carin, L., 2018, April. ''Zero-shot learning via class-conditioned deep generative models''. In Proceedings of the AAAI conference on artificial intelligence (Vol. 32, No. 1).

\bibitem{b61} Verma, V.K. and Rai, P., 2017. ''A simple exponential family framework for zero-shot learning''. In Machine Learning and Knowledge Discovery in Databases: European Conference, ECML PKDD 2017, Skopje, Macedonia, September 18–22, 2017, Proceedings, Part II 10 (pp. 792-808). Springer International Publishing.

\bibitem{b62} Mukherjee, T. and Hospedales, T., 2016, November. ''Gaussian visual-linguistic embedding for zero-shot recognition''. In Proceedings of the 2016 conference on empirical methods in natural language processing (pp. 912-918).

\bibitem{b63} Sun, X., Gu, J. and Sun, H., 2021. ''Research progress of zero-shot learning''. Applied Intelligence, 51, pp.3600-3614.

\bibitem{b64} Zhang, L., Sung, F., Liu, F., Xiang, T., Gong, S., Yang, Y. and Hospedales, T.M., 2017. ''Actor-critic sequence training for image captioning''. arXiv preprint arXiv:1706.09601.

\bibitem{b65} Zhang, Z. and Saligrama, V., 2015. ''Zero-shot learning via semantic similarity embedding''. In Proceedings of the IEEE international conference on computer vision (pp. 4166-4174).

\bibitem{b66} Liu, Y., Zhou, L., Bai, X., Huang, Y., Gu, L., Zhou, J. and Harada, T., 2021. ''Goal-oriented gaze estimation for zero-shot learning''. In Proceedings of the IEEE/CVF conference on computer vision and pattern recognition (pp. 3794-3803).

\bibitem{b67} Liu, B., Dong, Q. and Hu, Z., 2020, April. ''Zero-shot learning from adversarial feature residual to compact visual feature''. In Proceedings of the AAAI Conference on Artificial Intelligence (Vol. 34, No. 07, pp. 11547-11554).

\bibitem{b68} Palatucci, M., Pomerleau, D., Hinton, G.E. and Mitchell, T.M., 2009. ''Zero-shot learning with semantic output codes''. Advances in neural information processing systems, 22.

\bibitem{b69} Reed, S., Akata, Z., Yan, X., Logeswaran, L., Schiele, B. and Lee, H., 2016, June. ''Generative adversarial text to image synthesis''. In International conference on machine learning (pp. 1060-1069). PMLR.

\bibitem{b70} Liu, Y., Gao, X., Gao, Q., Han, J. and Shao, L., 2020. ''Label-activating framework for zero-shot learning''. Neural Networks, 121, pp.1-9.

\bibitem{b71} Bengio, Y., 2009. ''Learning deep architectures for AI''. Foundations and trends® in Machine Learning, 2(1), pp.1-127.

\bibitem{b72} ''Higher Algebra Abstract and Linear (11th ed.)'' Mapa, Sadhan Kumar. Sarat Book House. p. 36. ISBN 978-93-80663-24-1.

\bibitem{b73} Bora, A., Jalal, A., Price, E. and Dimakis, A.G., 2017, July. ''Compressed sensing using generative models''. In International Conference on Machine Learning (pp. 537-546). PMLR.

\bibitem{b74} ''Ch 2: Artificial Intelligence and Machine Learning for EDGE Computing11'' Tiwari, Ashish, Academic Press, 2022, Pages 23-32, ISBN 9780128240540

\bibitem{b75} Kothari, K., Khorashadizadeh, A., de Hoop, M. and Dokmanić, I., 2021, December. ''Trumpets: Injective flows for inference and inverse problems''. In Uncertainty in Artificial Intelligence (pp. 1269-1278). PMLR.

\bibitem{b76} Kynkäänniemi, T., Karras, T., Laine, S., Lehtinen, J. and Aila, T., 2019. ''Improved precision and recall metric for assessing generative models''. Advances in Neural Information Processing Systems, 32.

\bibitem{b77} Chen, S., Hong, Z., Liu, Y., Xie, G.S., Sun, B., Li, H., Peng, Q., Lu, K. and You, X., 2022, June. ''Transzero: Attribute-guided transformer for zero-shot learning''. In Proceedings of the AAAI Conference on Artificial Intelligence (Vol. 36, No. 1, pp. 330-338).

\bibitem{b78} Zhang, M., Wang, X., Shi, Y., Ren, S. and Wang, W., 2023. ''Zero-Shot Learning with Joint Generative Adversarial Networks''. Electronics, 12(10), p.2308.

\bibitem{b79} Cheng, D., Wang, G., Wang, B., Zhang, Q., Han, J. and Zhang, D., 2023. ''Hybrid routing transformer for zero-shot learning''. Pattern Recognition, 137, p.109270.

\bibitem{b80} Al Machot, F., Ullah, M. and Ullah, H., 2022. ''HFM: A Hybrid Feature Model Based on Conditional Auto Encoders for Zero-Shot Learning''. Journal of Imaging, 8(6), p.171.

\end{thebibliography}

\end{document}